\documentclass{article}

\usepackage{arxiv}

\usepackage[utf8]{inputenc} 
\usepackage[T1]{fontenc}    
\usepackage{hyperref}       
\usepackage{url}            
\usepackage{booktabs}       
\usepackage{amsfonts}       
\usepackage{nicefrac}       
\usepackage{microtype}      
\usepackage{lipsum}         
\usepackage{makecell}
\usepackage{graphicx}
\usepackage{multirow}

\title{A Clarifying Question Selection System from NTES\_ALONG in Convai3 Challenge}

\author{
  Wenjie Ou*, Yue Lin\\
  NetEase Games AI Lab \\
  \{ouwenjie, gzlinyue\}@corp.netease.com \\
}

\begin{document}
\maketitle

\begin{abstract}
This paper presents the winner solution system of NetEase Game AI Lab team for the ClariQ challenge at Search-oriented Conversational AI (SCAI) EMNLP workshop in 2020. The challenge asks for a complete conversational information retrieval system that can understand and generate clarification questions. We propose a clarifying question selection system that consists of response understanding, candidate question recalling and clarifying question ranking. We fine-tune a ELECTRA model to achieve a better understand of user's responses and use an enhanced BM25 model to recall the candidate questions. In clarifying question ranking stage, we reconstruct the training dataset and propose two models based on ELECTRA. Finally we ensemble the models by summing up their output probabilities and choose the question with the highest probability as the clarification question. Experiments show that our ensemble ranking model has a nice performance in the document relevance task and achieves the best recall@[20,30] metrics in question relevance task. And in multi-turn conversation evaluation in stage2, our system achieve the top score of all document relevance metrics.
\end{abstract}

\section{Introduction}
Search-oriented conversational systems aim to return an appropriate answer in response to the user requests. But some requests might be ambiguous and the systems can be enhanced by asking a good clarifying question. The main reserach challenges on clarifying questions for a conversational information retrival systems (ClariQ) are the following \cite{DBLP:journals/corr/abs-2009-11352}:

\begin{itemize}
\item When to ask clarifying questions during dialogues?
\item How to generate or select the clarifying questions?
\end{itemize}

\begin{figure}
  \centering
  \caption{System Architectures}
  \includegraphics[width=\textwidth]{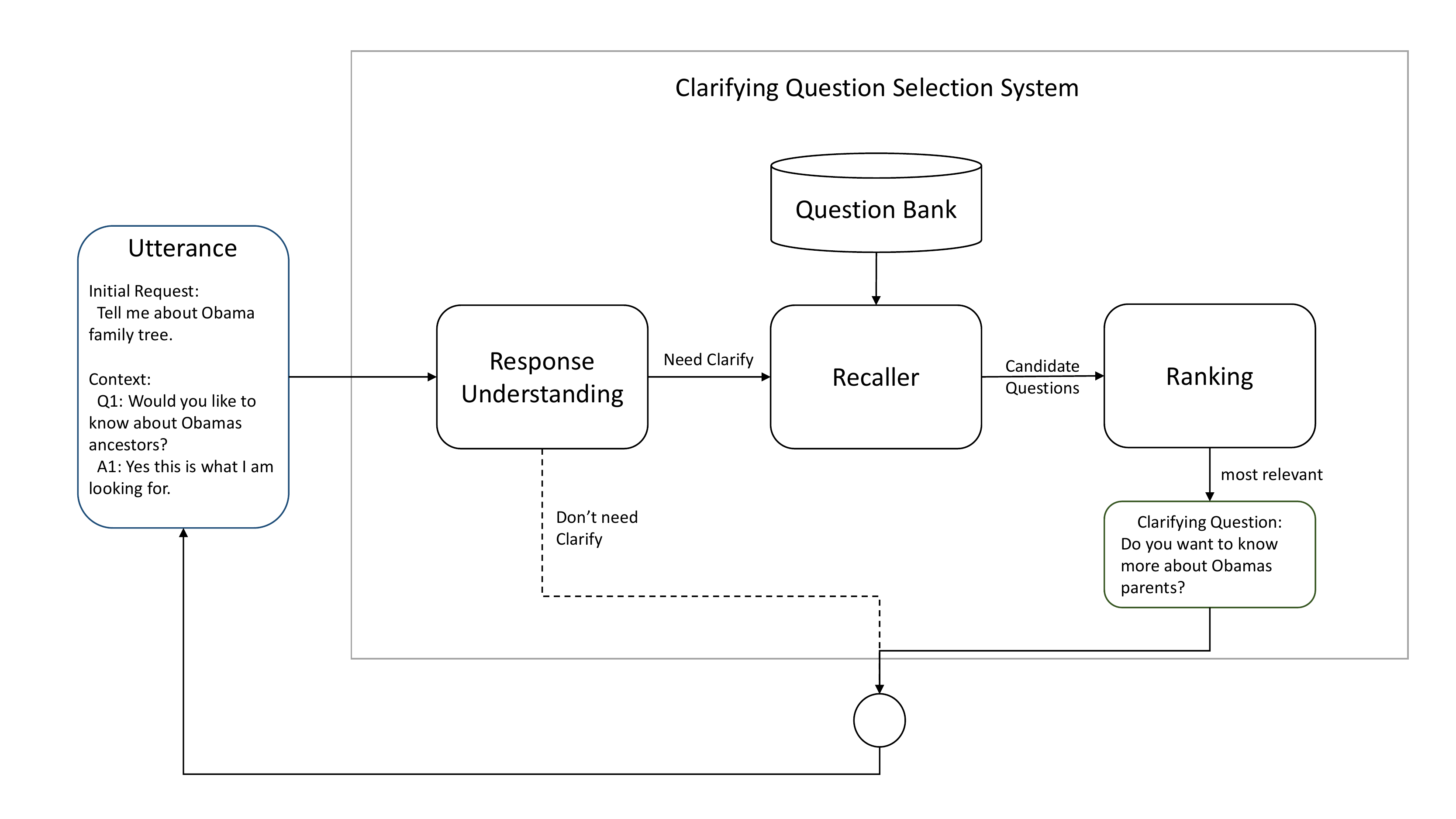}
  \label{fig:system}
\end{figure}

In this paper, we propose a clarifying question selection system that consists of response understanding, candidate question recalling and clarifying question ranking. The whole system architecture is shown in Figure \ref{fig:system}. And our system won the first place of the ConvAI3 challenge. Firstly we fine-tune a ELECTRA classification model to determine whether the user's responses need to be clarified. Then a candidate clarifying question recalling model is followed. We mainly focus on the clarifying question ranking task and treat it as a point-wise sequence classification task. We convert the related sequence pairs of the official dataset to the positive samples, and construct the negative samples by the BM25 model and randomly sampling approach, which are introduced in Section \ref{sec:clariq_data_preparation}. In Section \ref{sec:clariq_models} we describe the details of two models: a) a vallina ELECTRA transformer encoder, and b) a multi-task ELECTRA model. During inference, we sum up the output probabilities of these two models and take the highest probability question as the appropriate clarifying question. Experiments show that the ensemble model has a nice performance in the document relevance task and achieves the best recall@[20,30] metrics in question relevance task. And in multi-turn conversation evaluation in stage2, our system achieve the top score of all document relevance metrics.

\section{User Response Understanding}
\label{sec:user_response_understanding}
We solve the user response understanding task by constructing a corresponding dataset and fine-tune a sequence classification model based on ELECTRA discriminator \cite{DBLP:journals/corr/abs-2003-10555}.

\subsection{Data preparation}
\label{sec:understanding_data_preparation}
There are some system automatic evaluation metrics given by the official dataset, including MRR, P@[1,3,5,10,20], nDCG@[1,3,5,20]. These metrics are used to evaluate the quality of the selected clarifying question together with its corresponding answer. For the training data of the user response understanding model, we use P@5 metric to build the classification label, including $l_{need\_clarify}$ and $l_{no\_need\_clarify}$. The P@N decision support metric calculates the fraction of n recommendations that are good. So we label an ambiguous query to $l_{need\_clarify}$ if the P@5 score is 0, which means the query cannot be clearly described user’s need. On the other hand, label $l_{no\_need\_clarify}$ is set to the clear query while the P@5 score is greater than 0. Data statistics and some data samples are shown in Table \ref{tab:understanding_data}.

\begin{table}[]
\centering
\caption{Understanding Data Statistic}
\begin{tabular}{l|l|l|l|l}
\hline
\textbf{initial\_request}                                                  & \multicolumn{1}{c|}{\textbf{question}}                                                                          & \multicolumn{1}{c|}{\textbf{answer}}                                                                         & \multicolumn{1}{c|}{\textbf{label}} & \multicolumn{1}{c}{\textbf{\# of label}} \\ \hline
\begin{tabular}[c]{@{}l@{}}What is Fickle\\ Creek Farm\end{tabular}        & \begin{tabular}[c]{@{}l@{}}are you looking for\\ information about staying\\  at fickle creek farm\end{tabular} & no                                                                                                           & $l_{need\_clarify}$               & 4,895                                    \\ \hline
\begin{tabular}[c]{@{}l@{}}Tell me about\\ Obama family tree.\end{tabular} & \begin{tabular}[c]{@{}l@{}}would you like to know\\ about obamas ancestors\end{tabular}                         & \begin{tabular}[c]{@{}l@{}}yes particualarly\\ information about his\\ parents and grandparents\end{tabular} & $l_{no\_need\_clarify}$            & 3,671                                    \\ \hline
\end{tabular}
\label{tab:understanding_data}
\end{table}

\subsection{Model}
\label{sec:understanding_model}
We utilize ELECTRA, a transformer based attentive neural architecture. The pretrained large ELECTRA models have shown strong performance for the sequence classification \cite{DBLP:journals/corr/abs-2003-10555}. For fine-tuning the pretrained ELECTRA model on the previous dataset, we use the HuggingFace's Transformers library \cite{DBLP:journals/corr/abs-1910-03771}. The model is optimized 10 epochs with a batch size of 24 and a learning rate of 5e-6. The large ELECTRA architecture consists of 24-layer self-attention heads (1024 dimensional states and 16 attention heads) and a 2-layer fully connected network. The model architecture is the same as Figure \ref{fig:electra}. In the prediction stage, the understanding model takes the last clarifying question and user answer as input and determine if they need to be clarified (with label $l_{need\_clarify}$) or not (with label $l_{no\_need\_clarify}$).

\section{Candidate Question Recalling}
\label{sec:bm25_recaller}
The official BM25 baseline information retrieval model \cite{Manning:2135372} ranks all the questions from the question bank based on the query terms appearing in each question. For enhancing the BM25 model, we build the retrieval indexes of each question with not only its terms but also the terms from its related initial requests, answers and topic descriptions in official single turn training dataset.

Since we found that the shorter the question, the more general, all the questions which not appear in the training set were sorted ascendingly by the number of their terms, denoted as $Q_{na}$. For each user initial request, the candidate clarifying questions will be recalled not only by the enhanced BM25 model but also from $Q_{na}$. In our system setting, we recall totally 200 candidate questions for each utterance, including with top 100 most related questions from BM25 model and top 100 shortest questions from $Q_{na}$.

\section{Clarifying Question Ranking}
\label{sec:clariq_ranking}
\subsection{Data preparation}
\label{sec:clariq_data_preparation}
We convert this task to a sentence pair binary classification task. The positive samples are pairs contained the asked clarifying question and the initial request from the official training set. The negative samples are constructed by negative sampling. We try two methods to build the negative samples. One method is to randomly sample from all the irrelevant questions. Another method is to use the BM25 model for sampling, which is mentioned in Section \ref{sec:bm25_recaller}. The difference from the previous recalling model is that, for every initial request, we recall the top 200 related questions by BM25 model and randomly sample 300 questions from the question bank. Finally the size of our training dataset is 95,762 samples consisting of 2,600 positives and 93,162 negatives. For the development dataset in training stage, we use the same approach to build 25,603 samples including 681 positives and 24,922 negatives.

We also construct some document relevant scores of positive samples, by searching the MRR100 and nDCG@3 scores from the official given evaluation file, and zero for negative samples. Table \ref{tab:construced_data} shows some training samples in our training dataset and Table \ref{tab:data_statistic} shows the data statistics.

\begin{table}[]
\caption{constructed training data}
\begin{center}
\begin{tabular}{l|l|l|l|l}
\hline
\textbf{initial\_request}        & \textbf{question}                                                      & \textbf{label} & \textbf{MRR100} & \textbf{nDCG@3} \\ \hline
Tell me about Obama family tree. & \makecell[l]{do you want to know more about obamas \\ parents}                          & 1              & 0.5             & 0.3333       \\ \hline
Tell me about Obama family tree. & \makecell[l]{are you referring to the time magazine \\ essay}                           & 1              & 0.6667        & 0.4115       \\ \hline
Tell me about Obama family tree. & \makecell[l]{what would you like to know about his \\ family (\textit{recall by bm25})} & 0              & 0               & 0              \\ \hline
Tell me about Obama family tree. & \makecell[l]{do you want a one pot recipe \\ (\textit{randomly sample})}                         & 0              & 0               & 0              \\ \hline
\end{tabular}
\end{center}
\label{tab:construced_data}
\end{table}

\begin{table}[]
\caption{data statistic}
\begin{center}
\begin{tabular}{l|l}
\hline
\textbf{Feature}                          & \textbf{Value \#}       \\ \hline
\# of train / dev / test topics           & 187 / 50 / 62           \\ \hline
\# of total question                      & 3,929                   \\ \hline
\# of training data / positive / negative & 95,762 / 2,600 / 93,162 \\ \hline
\# of dev data / positive / negative      & 25,603 / 681 / 24,922   \\ \hline
\end{tabular}
\end{center}
\label{tab:data_statistic}
\end{table}

\subsection{Models}
\label{sec:clariq_models}
In this sentence ranking task, we fine-tune two sentence classification models. One is the vanilla ELECTRA-large-Discriminator, the other is a multi-task model. We will introduce more details in the next section \ref{sec:electra} and section \ref{sec:multi_task_model}. The fine-tuning details and the prediction details are introduced in section \ref{sec:finetuning_details} and \ref{sec:inference_details}.

\subsubsection{ELECTRA}
\label{sec:electra}

Referring to BERT's sentence pair data construction method \cite{DBLP:conf/naacl/DevlinCLT19}, we add a "[SEP]" token at the end of the initial\_request value and the question value. After packing together the sentence pairs, we add a "[CLS]" token before the first word. Finally we take this new single sequence as the input of the model.

\begin{figure}
  \centering
  \caption{Model Architectures}
  \begin{minipage}[t]{0.48\textwidth}
  \includegraphics[width=\textwidth]{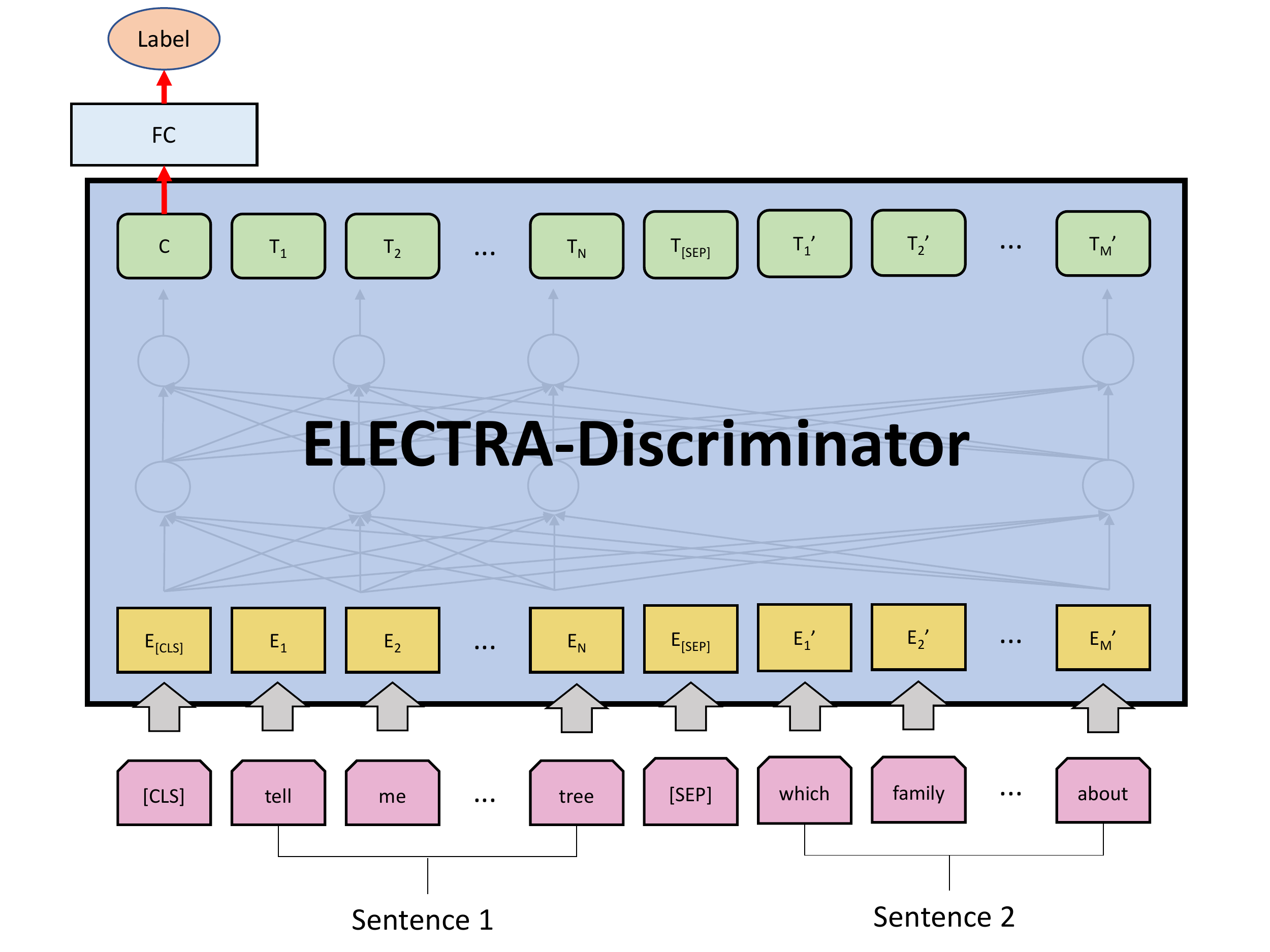}
  \label{fig:electra}
  \end{minipage}
  \begin{minipage}[t]{0.48\textwidth}
  \includegraphics[width=\textwidth]{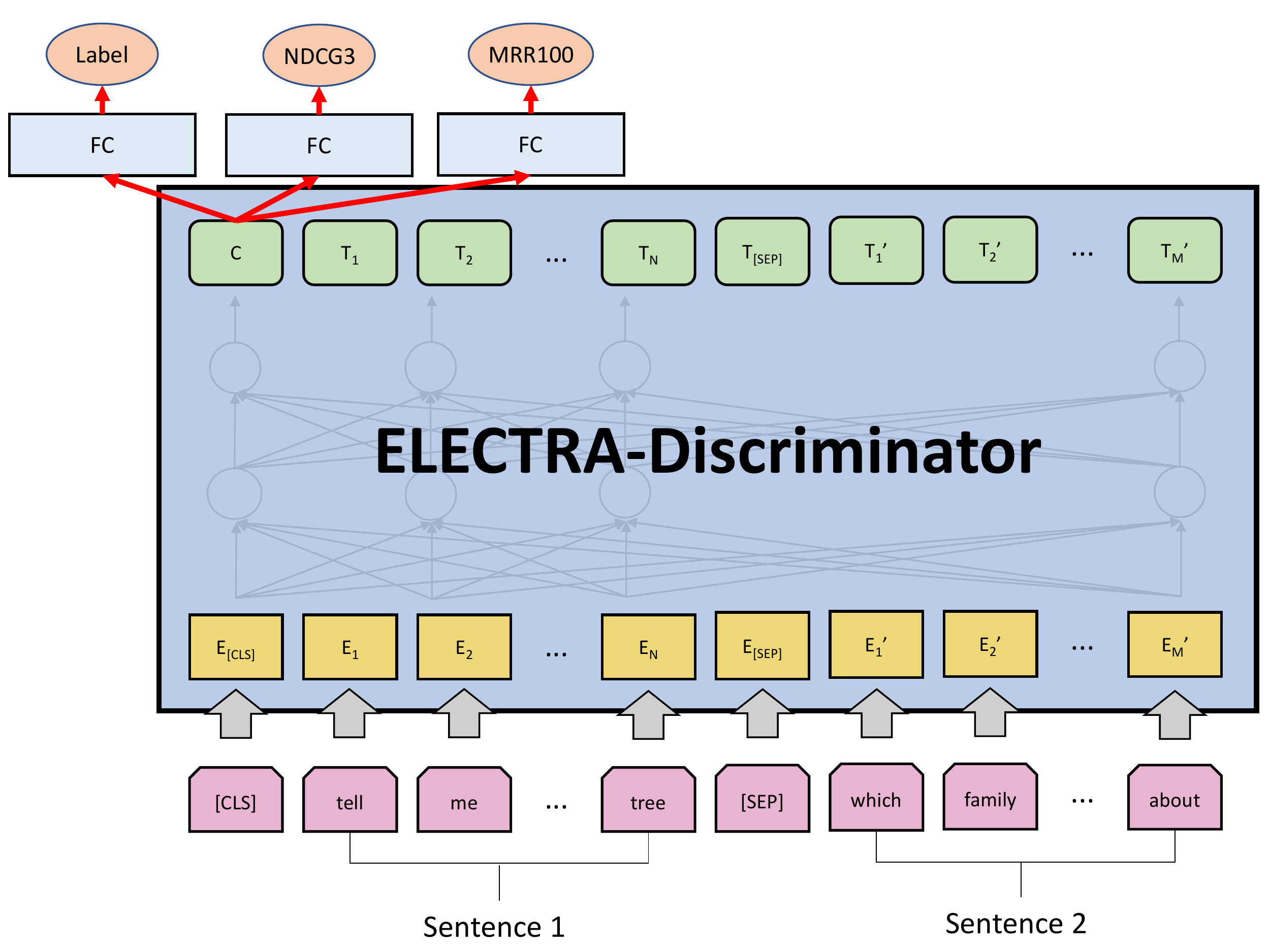}
  \label{fig:electra_multitask}
  \end{minipage}
\end{figure}

Our sentence pair classification model is a multilayer Transformer encoder based on the ELECTRA \cite{DBLP:journals/corr/abs-2003-10555}.  We use a large size model configuration and more details is in Section \ref{sec:finetuning_details}. Figure \ref{fig:electra} shows the architecture of our sequence pair classification model.

In fine-tuning stage, we take the final hidden state $h_{[cls]}$ to represent the input data, which is the transformer output of the first token of the sequence. Then $h_{[cls]}$ will be fed into a two-layer fully connected network activated by the GLEU activation function \cite{DBLP:journals/corr/HendrycksG16} and output the sentence representation $h_s = FC(h_{[cls]})$.  Finally the label probabilities are computed with a standard $softmax$, $p = softmax(h_s)$. All of the parameters of the electra-discriminator and FC layer are fine-tuned jointly to minimize the negative log-probability of the correct label.

\subsubsection{Multi-task Model}
\label{sec:multi_task_model}

For the input data of the model, we apply the same data construction method in Section \ref{sec:electra}. While using the classification label, we added two numerical values as the additional regression tasks to fine-tune the model, detailed in Figure \ref{fig:electra_multitask}.

The sequence encoder of multi-task model is also the same as the one in Section \ref{sec:electra},  which outputs the sentence representation $h_{[cls]}$. The classification task model is also the same as the previous model. For the two regression tasks, each output layer is constructed by a two-layer full connected network and a $sigmoid$ activation layer. Finally the fine-tuning of the parameters of the multi-task model is done by optimizing a combination of a binary classification cross entropy loss $Loss_{bce}$ and two MSE regression loss $MSE_{ndcg}$ and $MSE_{mrr}$.

\begin{equation}
    Loss_{total} = Loss_{bce} + MSE_{ndcg} + MSE_{mrr}
\end{equation}

\subsubsection{Fine-tuning Details}
\label{sec:finetuning_details}

We use a 24-layer large ELECTRA-discriminator as the sequence encoder with the self-attention heads (1024 dimensional states and 16 attention heads).  The electra-discriminator is based on a recently published PyTorch adaptation by the HuggingFace's Transformer library \cite{DBLP:journals/corr/abs-1910-03771}. All the parameters of ELECTRA are initialized from the pretrained electra-large-discriminator model presented by google team \footnote{\url{https://huggingface.co/google/electra-large-discriminator}}.

We fine-tune the model with a batch size of 24 sequences having a max of 256 tokens for 1 epoch, which is approximately 4000 steps over our constructed training dataset.  We used the training framework implemented by Huggingface's Transformers, setting a learning rate of 5e-6. The learning rate was linearly decayed to zero over the course of the training. Fine-tuning the model takes about 2h on four GeForce GTX 1080Ti GPUs.

\subsubsection{Inference Details}
\label{sec:inference_details}
As for the stage 1 of the ClariQ challenge, we select the top 30 highest probability questions for each initial request from an ensemble model, which sum up the predicted probabilities of the previous two models. But as for our system, we select the most relevant clarifying question from the recalling candidate question set and return it to the user.
It is worth mentioning that these two models are trained on the single turn dataset and take the combinations of initial request and conversation context as input in the inference stage. For more details to the system input, the clarifying question will be added into the initial requests as the input if there is a "yes" in the corresponding answer. If there is a "no" in the user's answer, we won't add anything to the initial request. In other cases, the system input is the initial request together with the user's answer.

\subsection{Experiments}

\begin{table}[]
\centering
\caption{Results of Document Relevance Task}
\begin{tabular}{l|c c c c|c c c c}
\hline
\multicolumn{1}{c|}{\multirow{2}{*}{\textbf{model}}} & \multicolumn{4}{c|}{\textbf{devset}}                               & \multicolumn{4}{c}{\textbf{testset}}                                 \\ \cline{2-9} 
\multicolumn{1}{c|}{}                       & \textbf{MRR}    & \textbf{P@1} & \textbf{NDCG@3} & \textbf{NDCG@5} & \textbf{MRR}    & \textbf{P@1}    & \textbf{NDCG@3} & \textbf{NDCG@5} \\ \hline
bm25 baseline                                & 0.3096          & 0.2313       & 0.1608          & 0.1530          & 0.3134          & 0.2193          & 0.1151          & 0.1061          \\
bert reranker                                & 0.3453          & 0.2563       & 0.1824          & 0.1744          & 0.2553          & 0.1784          & 0.0892          & 0.0818          \\ \hline
electra                                      & 0.3548          & 0.2436       & 0.2013          & 0.1921          & -               & -               & -               & -               \\
electra multitask                            & 0.3520          & 0.2688       & 0.2033          & 0.1925          & 0.3006          & 0.2230          & 0.1031          & 0.097           \\
\textbf{ensemble model}                      & \textbf{0.3761} & \textbf{0.3} & \textbf{0.2113} & \textbf{0.1955} & \textbf{0.3140} & \textbf{0.2379} & \textbf{0.1229} & \textbf{0.1097} \\ \hline
\end{tabular}
\label{tab:document_relevance}
\end{table}

\begin{table}[]
\centering
\caption{Results of Question Relevance Task}
\begin{tabular}{l|c c c c|c c c c}
\hline
\multicolumn{1}{c|}{\multirow{2}{*}{\textbf{model}}} & \multicolumn{4}{c|}{\textbf{dev}}                                     & \multicolumn{4}{c}{\textbf{test}}                                  \\ \cline{2-9} 
\multicolumn{1}{c|}{}                                & \textbf{R@5}    & \textbf{R@10}   & \textbf{R@20}   & \textbf{R@30}   & \textbf{R@5}    & \textbf{R@10} & \textbf{R@20}   & \textbf{R@30}   \\ \hline
bm25 baseline                                         & 0.3245          & 0.5638          & 0.6675          & 0.6913          & 0.3170          & 0.5705        & 0.7292          & 0.7682          \\
bert rerank                                           & 0.3475          & 0.6122          & 0.6913          & 0.6913          & 0.3444          & 0.6062        & 0.7585          & 0.7682          \\ \hline
electra                                               & 0.3604          & 0.6618          & 0.8368          & 0.8632          & -               & -             & -               & -               \\
\textbf{electra multitask}                            & \textbf{0.3648} & \textbf{0.6753} & \textbf{0.8510} & 0.8744          & \textbf{0.3414} & \textbf{0.6351}        & 0.8316          & 0.8721          \\
\textbf{ensemble model}                               & 0.3604          & 0.6749          & 0.8478          & \textbf{0.8761} & 0.3404          & 0.6329        & \textbf{0.8335} & \textbf{0.8744} \\ \hline
\end{tabular}
\label{tab:question_relevance}
\end{table}

\subsubsection{Evaluation Metric}

We send the top 30 highest probability clarifying questions of the user requests from the single turn test dataset to the task organizer for automatic evaluation. The evaluation metrics of document relevance task here are MRR, P@3 and nDCG@[3,5]. The selected clarifying question and its corresponding answer are added to the original request, which is then used to retrieve documents from the collection. These metrics are evaluated how much the question and its answer affect the performance of document retrieval.For the question relevance task, the models are evaluated in terms of Recall@[5, 10, 20, 30]. More details about automatic evaluation metrics can be found in the challenge description \cite{DBLP:journals/corr/abs-2009-11352}.

In the multi-turn conversations evaluation in stage2, our system achieve the top scores in all metrics, which is shown in \ref{tab:system_evaluation}.

\subsubsection{Results}

The results of our experiments are summarized in Table \ref{tab:document_relevance} and Table \ref{tab:question_relevance}. We utilize the official BM25 baseline and BERT reranker as the benchmark here for comparison. For our models, we compared three different configurations: ELECTRA, multi-task model based on ELECTRA, and the ensemble of these two models. The final leaderboard can be found at the official challenge website \footnote{\url{https://github.com/DeepPavlov/convai}}.

Table \ref{tab:document_relevance} shows that the ensemble model outperforms in both development dataset and test dataset in the document relevance task. As for question relevance task whose result is shown in Table \ref{tab:question_relevance}, ELECTRA-multitask model achieves the best recall@[5,10] and the ensemble model achieves the best recall@[20,30] in single turn test dataset.

Table \ref{tab:system_result} shows some examples of the system simulation. Our system chooses the question which asks for the user's purpose in the first case and selects a more detailed clarifying question in the second case. In the third case, our system asks a further question according to the conversation context. And in the fourth case, the system asks the question different from the last question if the user answers no. When the user's answer gives more details, which is shown in the fifth case, our system no longer asks questions.

\begin{table}[]
\centering
\caption{Results of Multi-turn Conversations Evaluation}
\begin{tabular}{l | l | c c c c}
\hline
\textbf{Rank} & \textbf{Creator}     & \textbf{MRR}    & \textbf{P@1}    & \textbf{nDCG@3} & \textbf{nDCG@5} \\ \hline
\textbf{1}    & \textbf{NTES\_ALONG} & \textbf{0.1798} & \textbf{0.1161} & \textbf{0.0553} & \textbf{0.0536} \\ \hline
2             & TAL ML               & 0.1669          & 0.1067          & 0.0522          & 0.0494         \\ \hline
\end{tabular}
\label{tab:system_evaluation}
\end{table}

\begin{table}[]
\centering
\caption{Results of System Simulation}
\begin{tabular}{l}
\hline
\begin{tabular}[c]{@{}l@{}}initial request   : I want to know about appraisals.\\ conversation context :\\  \ \  None\\ clarifying question : What kind of appraisal are you looking for ?\end{tabular}                                                                                                                   \\ \hline
\begin{tabular}[c]{@{}l@{}}initial request   : How to cure angular cheilitis ?\\ conversation context :\\  \ \  None\\ clarifying question : Are you interested in home remedies for angular cheilitis ?\end{tabular}                                                                                                     \\ \hline
\begin{tabular}[c]{@{}l@{}}initial request   : I want to know about appraisals.\\ conversation context :\\ \ \  question  : What kind of appraisal are you looking for ?\\ \ \  answer   : I need information about antique appraisals.\\ clarifying question : Would you like to find appraisers near you ?\end{tabular} \\ \hline
\begin{tabular}[c]{@{}l@{}}initial request   : Where can I buy pressure washers ?\\ conversation context :\\ \ \  question  : Are you wondering what a pressure washer is ?\\ \ \  answer   : No.\\ clarifying question : Are you looking for a place to buy pressure washer parts ?\end{tabular}                         \\ \hline
\begin{tabular}[c]{@{}l@{}}initial request   : Find information on Hoboken .\\ conversation context :\\  \ \  question  : Are you looking for an apartment in hoboken ?\\  \ \  answer   : No, I would like to find restaurants there.\\ clarifying question : None (means the request is clear)\end{tabular}               \\ \hline
\end{tabular}
\label{tab:system_result}
\end{table}

\section{Conclusions and Future Work}

For the challenge of clarifying questions for conversational information retrieval systems (ClariQ), we present a clarifying question selection system which is a pipeline approach combining with response understanding, candidate question recalling and clarifying question ranking. We firstly fine-tune a ELECTRA classification model to determine which utterance needs to be clarified. For the ambiguous requests, our system recalls the candidate questions by an enhanced BM25 method and select the most related from them by an ensemble ranking model. Especially for the ensemble ranking model, we fine-tune an ELECTRA model and a multi-task learning ELECTRA model with our ranking dataset. Compared with the benchmark models, our models have showed competitive performances in document relevance task. And in question relevance task, our ensemble model ranks 1st among all participant models. In practical application, our system tends to return the clear and coherent questions to clarify the user’s query. And our system won the first place of the ConvAI3 challenge in EMNLP 2020 SCAI workshop.

For future work, we plan to investigate whether applying different interaction linking structures and tune the hyper-parameters extensively can further improve the system’s performance. In addition, we will apply our framework on various dataset from different domains to evaluate its generalize ability and robustness.

\bibliographystyle{unsrt}  

\bibliography{references}

\begin{thebibliography}{1}

\bibitem{DBLP:journals/corr/abs-2009-11352}
Mohammad Aliannejadi, Julia Kiseleva, Aleksandr Chuklin, Jeff Dalton, and
  Mikhail~S. Burtsev.
\newblock Convai3: Generating clarifying questions for open-domain dialogue
  systems (clariq).
\newblock {\em CoRR}, abs/2009.11352, 2020.

\bibitem{DBLP:journals/corr/abs-2003-10555}
Kevin Clark, Minh{-}Thang Luong, Quoc~V. Le, and Christopher~D. Manning.
\newblock {ELECTRA:} pre-training text encoders as discriminators rather than
  generators.
\newblock {\em CoRR}, abs/2003.10555, 2020.

\bibitem{DBLP:journals/corr/abs-1910-03771}
Thomas Wolf, Lysandre Debut, Victor Sanh, Julien Chaumond, Clement Delangue,
  Anthony Moi, Pierric Cistac, Tim Rault, R{\'{e}}mi Louf, Morgan Funtowicz,
  and Jamie Brew.
\newblock Huggingface's transformers: State-of-the-art natural language
  processing.
\newblock {\em CoRR}, abs/1910.03771, 2019.

\bibitem{Manning:2135372}
Christopher~D Manning, Prabhakar Raghavan, and Hinrich Schütze.
\newblock {\em {Introduction to information retrieval}}.
\newblock Cambridge University Press, Cambridge, 2008.

\bibitem{DBLP:conf/naacl/DevlinCLT19}
Jacob Devlin, Ming{-}Wei Chang, Kenton Lee, and Kristina Toutanova.
\newblock {BERT:} pre-training of deep bidirectional transformers for language
  understanding.
\newblock In Jill Burstein, Christy Doran, and Thamar Solorio, editors, {\em
  Proceedings of the 2019 Conference of the North American Chapter of the
  Association for Computational Linguistics: Human Language Technologies,
  {NAACL-HLT} 2019, Minneapolis, MN, USA, June 2-7, 2019, Volume 1 (Long and
  Short Papers)}, pages 4171--4186. Association for Computational Linguistics,
  2019.

\bibitem{DBLP:journals/corr/HendrycksG16}
Dan Hendrycks and Kevin Gimpel.
\newblock Bridging nonlinearities and stochastic regularizers with gaussian
  error linear units.
\newblock {\em CoRR}, abs/1606.08415, 2016.

\end{thebibliography}

\end{document}